\pdfoutput=1

\documentclass[11pt]{article}

\usepackage[final]{acl}

\usepackage{txfonts}
\usepackage{latexsym}
\usepackage{array} 

\usepackage[T1]{fontenc}

\usepackage[utf8]{inputenc}

\usepackage{microtype}

\usepackage{inconsolata}

\usepackage{graphicx}
\usepackage{float}
\usepackage{afterpage}

\title{Historical Ink: 19th Century Latin American Spanish Newspaper\\ Corpus with LLM OCR Correction}

\author{
  Laura Manrique-Gómez$^1$ \quad Tony Montes$^2$ \quad Arturo Rodríguez-Herrera$^3$ \quad Rubén Manrique$^2$  \\
  $^1$ History and Geography Department, Universidad de los Andes, Bogotá D.C. \\
  $^2$ Systems and Computing Engineering Department, Universidad de los Andes, Bogotá D.C. \\  
  $^3$ Civil and Environmental Engineering Department, Rice University, Houston TX \\\ \\
  \texttt{\{l.manriqueg, t.montes, rf.manrique\}@uniandes.edu.co} \\
  \texttt{da.rodriguezh@rice.edu}\\
}

\begin{document}
\maketitle
\begin{abstract}

This paper presents two significant contributions: First, it introduces a novel dataset of 19th-century Latin American newspaper texts, addressing a critical gap in specialized corpora for historical and linguistic analysis in this region. Second, it develops a flexible framework that utilizes a Large Language Model for OCR error correction and linguistic surface form detection in digitized corpora. This semi-automated framework is adaptable to various contexts and datasets and is applied to the newly created dataset.

\end{abstract}

\section{Introduction}
The computational processing of historical newspaper texts is crucial due to the valuable information these texts contain about political, economic, and cultural history. Over the past three decades, Digital Humanities has driven extensive digitization efforts, resulting in numerous curated digital collections \cite{Berry_Fagerjord_2017, Dobson_2019}. However, converting these images into machine-readable texts remains challenging, particularly in achieving accurate transcription. A primary challenge is the accuracy of OCR technology, especially with the extremely diverse newspaper layouts, materially degraded documents, and non-standardized fonts typical of historical texts. Traditional OCR methods often produce errors that complicate subsequent analysis.

To address these challenges, we employed GPT-4o-mini \cite{gpt-4o-mini}, a Large Language Model (LLM), within a pipeline for OCR error correction. While the LLM is capable of fixing OCR-related errors that traditional systems often miss \cite{Post-OCR-Correction}, our pipeline also detects and classifies potential hallucinations to avoid further issues and streamline the process. Additionally, it contributes by identifying surface forms—specific word occurrences—within the dataset.

\subsection{Related Work}

The "Chronicling America" initiative marks a significant advancement in the digitization of historical newspaper materials \cite{Chronicling_America}. Another major effort, is the "Atlas - Oceanic Exchanges" collection, which traces global information networks in 19th-century newspaper materials \cite{Oceanic_Exchanges_2021}. Similarly, “Viral Texts: Mapping Networks of Reprinting in 19th-Century Newspapers and Magazines” \cite{Cordell_Smith_2022} explores the culture of reprinting in the U.S. before the Civil War, while the European “Project Impresso: Media Monitoring the Past” \cite{Impresso_2023} addresses the OCR challenges specific to English and Germanic languages.

Despite these advancements, historical newspapers are scarcely digitized in the Global South \cite{LeBlanc2024}. Consequently, a gap remains in specialized corpora for 19th-century Latin American newspapers, limiting the study of the region's unique historical and linguistic features. Our research addresses this gap by introducing a new dataset of Latin American newspaper texts in old Spanish. This dataset was post-processed with LLM models for addressing OCR errors and distinguishing them from historical linguistic surface forms\footnote{The dataset is available at \url{https://huggingface.co/datasets/Flaglab/latam-xix} in its three versions: "original", "cleaned", and "corrected"}.

ICDAR post-OCR correction competitions in 2017 and 2019 \cite{8270163, 8978127} presented interesting solutions to error detection and correction in 10 European languages, such as Clova AI model based on multi-lingual BERT. Similarly,  \citet{10.1145/3383583.3398605} achieved comparable results by initializing embeddings with popular static embeddings such as GloVe \cite{pennington-etal-2014-glove}. In another approach, \citet{veninga} examined the fine-tuning of ByT5, a character-level LLM, emphasizing the importance of preprocessing and context length optimization. This results aligns with earlier studies on character-level models, such as \citet{Amrhein2018}, which demonstrated the potential of character-based sequence-to-sequence models in improving OCR correction. 

The application of LLMs for post-OCR correction has gained traction, especially in improving the accuracy of digitized historical texts. Early work by \citet{Nguyen2021} laid the foundation by categorizing post-OCR correction methods, highlighting the challenges associated with isolated-word and context-dependent approaches. As discussed by \citet{thomas-etal-2024-leveraging}, the introduction of Transformers' architecture leads to state-of-the-art performance in various text correction tasks and also presents a new baseline for post-OCR correction.

\citet{Post-OCR-Correction} builds on this foundation by addressing the persistent issue of OCR quality in cultural heritage texts. They propose that LLMs can significantly enhance correction accuracy through context-aware processing, although challenges like hallucinations and language switching remain. More recent work by \citet{thomas-etal-2024-leveraging} demonstrates the superiority of a prompt-based approach using Llama 2 over traditional models like BART \cite{bart}, reducing character error rates (CER) by over 54\%. These findings are consistent with those of \citet{Soper2021}, who reported comparable improvements using fine-tuned BART models. These studies highlight the evolution from traditional correction methods to LLM-based approaches. Nevertheless, further studies are needed to test correction methods in historical documents containing linguistic and regional variants.

\section{Sourcing}

The dataset was initially compiled from Colombian digital newspaper archives. The primary focus was on publications that included cartoons or illustrations, which were intended for subsequent multimodal modeling. This review also extended to the physical collections on-site, as only approximately 50\% of the physical collection had been digitized. Through this process, 64 newspaper titles were identified, representing 7\% of the total 1,655 publications in the collections. This first iteration resulted in a dataset consisting of 4,032 pages of scanned pages of newspapers, primarily from Nueva Granada—a former country encompassing Colombia, Panama, Venezuela, and Ecuador—.

\begin{figure}[t]
\centering
  \includegraphics[width=\linewidth]{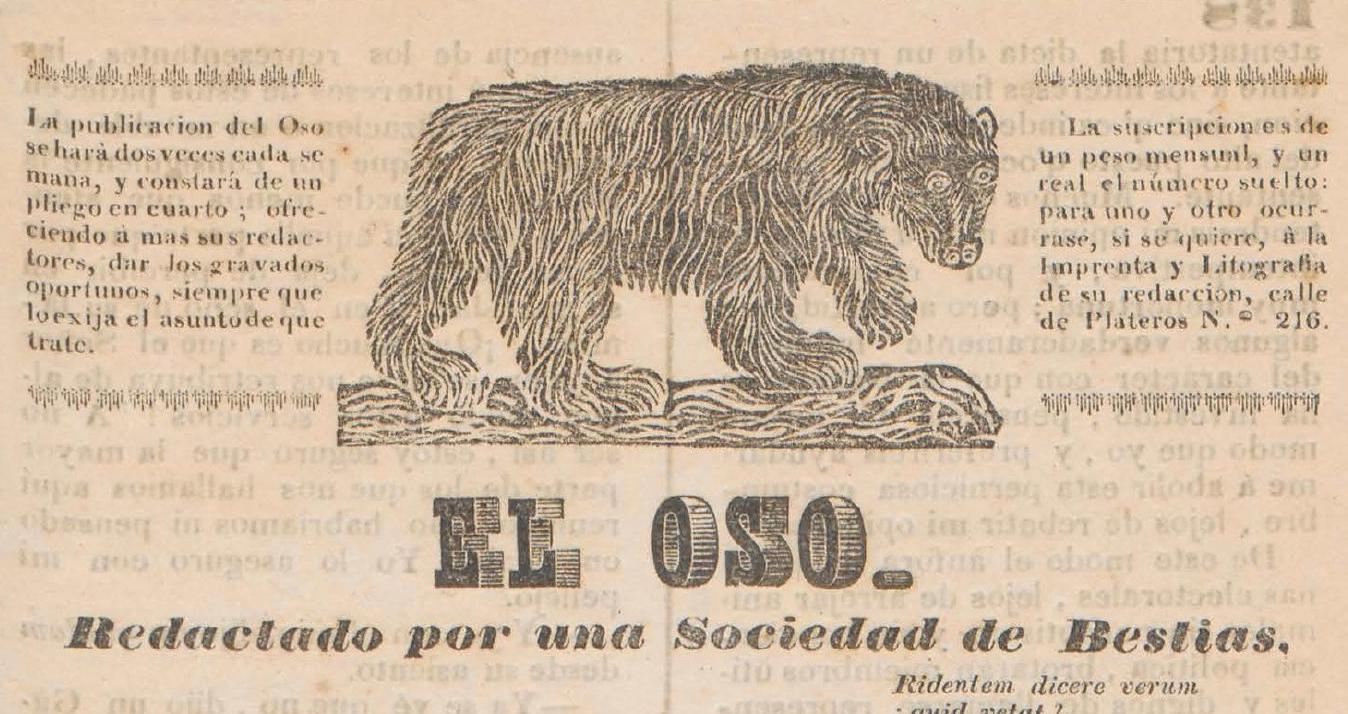}
  \caption{El Oso, Peru. An example of a scanned newspaper image. The corresponding OCR-extracted text and the corrected version can be found in Appendix \ref{sec:appendix1}, for reference.}
  \label{fig:example}
\end{figure}

A second iteration completed the revision of 3,038 digitized newspapers of 58 digital collections across Mexico, Argentina, Colombia, Peru, Chile, Panama, Venezuela, Uruguay, Bolivia, Cuba, and Ecuador as shown in Table \ref{tab:overview-country}
. Some countries, such as Bolivia, Cuba, and Venezuela have very limited or no web collections, resulting in their underrepresentation or absence from the final dataset. Additionally, some newspapers were printed in Europe due to lower costs; in some cases, printing outsourcing was utilized. The final dataset comprises 197 newspaper titles and 23,522 pages of scanned images, primarily from Mexico City (Mexico is the only country that has digitized its entire collection), but also includes publications from other Latin American cities, such as Buenos Aires, Lima, Bogota, and Santiago de Chile. An example of a newspaper image can be observed in Figure \ref{fig:example}.

Originally, the Latin American 19-century newspaper dataset consists of scanned images. These images were processed using a layout model, followed by an OCR service. The layout model was specifically trained using data from annotated newspapers available in Roboflow \citet{dataset1,dataset2,dataset3,dataset4}. These datasets were merged into a single dataset (CD) consisting of 1368 images of newspapers annotated for binary layout classification: images and texts. The CD dataset includes 10\% of images from our newspapers dataset, labeled by hand, and it was enriched with data augmentation for shear and rotation. These techniques help to increase the model's performance in images with scanning errors. 

The CD dataset was used to train an image recognition model from Azure Cognitive Services\footnote{Model available through Azure cloud services at \url{https://learn.microsoft.com/en-us/azure/ai-services/computer-vision/}}, which can extract the images in the newspaper page and extract the text through the OCR. The model's performance scored MAP@75 of 87.0\%, resulting in a collection of annotations and coordinates for both text and images. These coordinates were used to crop the original image, and then process it with the OCR model. Once the OCR results were obtained, we merged the processed text with the images, creating a dataset that contains the newspaper images and their associated text. From a sample of 2,500 transcribed texts, each containing 1,000 characters, manual supervision revealed that 8.5\% were unreadable. The remaining texts contained multiple transcription errors, primarily due to the artisanal printing techniques and the grammatical and lexical variations of the era. These errors significantly impacted readability, introducing bias when using the texts as input for NLP-LLM models.

\section{Processing}

The dataset includes samples of newspapers that were either handwritten or produced using early carving machines. Over time, these machines would wear out, leading to text features that were easily confused with backward accent marks, unwanted punctuations, or misplaced characters between words. Such misreadings disrupted the continuity of the text without adding any semantic meaning.

Detecting these errors automatically poses a challenge due to the linguistic shifts between modern and 19th-century Spanish. OCR models trained on such historical texts are lacking, especially considering the semantic and orthographic changes over time. For instance, what might appear as an OCR error could instead be a historical surface form of a word; for example, the conjunction "y" (and) was often written as "I".

\begin{figure*}[t]
\centering
  \includegraphics[width=\linewidth]{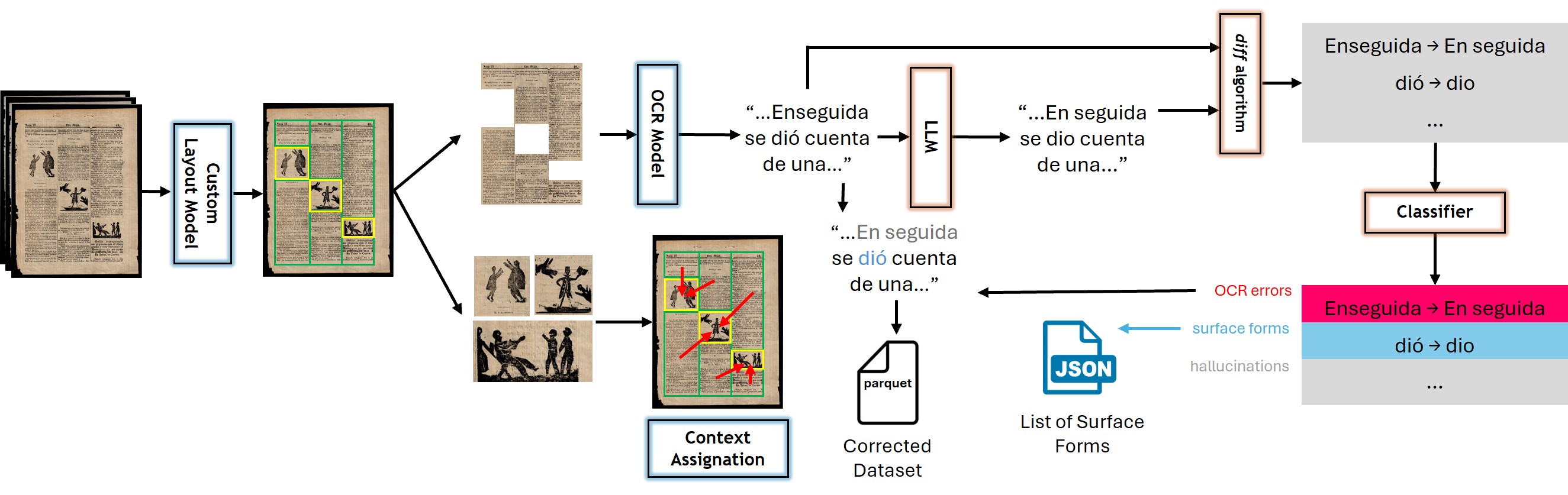}
  \caption{Overview of the full methodology pipeline. The {\color{blue} blue} components correspond to the Layout+OCR stage to get to digitized text, and the {\color{orange} orange} components correspond to the Post-OCR LLM Correction stage. The two outputs of the pipeline are the LatamXIX \textbf{Corrected Dataset} and the \textbf{List of Surface Forms}. The \textit{Custom Layout Model} also extracts the images of the newspaper which are then assigned to the related texts (context). The final version of the text has the OCR errors corrected but not the surface forms, as they are part of the language.}
  \label{fig:pipeline}
\end{figure*}

Additionally, some texts were completely unintelligible for OCR, and challenging for humans to interpret, due to the fonts used in certain newspapers. The varied layouts of these newspapers also resulted in texts filled with scores or numbers, or in some cases, samples containing only chapter titles or numbering (e.g., "III IV V"), which added noise to the dataset. A general overview of the pipeline from the source until the final post-processed, is observed in Figure \ref{fig:pipeline}.

\subsection{Cleaning and filtering}

Some of the most common cleaning steps for text data include removing duplicates and noisy data, which are particularly crucial for subsequent analysis. In this case, $3.08\%$ of rows were removed due to duplicates or empty texts. Additionally, $1.74\%$ of rows were filtered out where over $50\%$ of the characters were non-alphabetic, as these rows are more likely to be noise than useful content. Rows with four or fewer tokens were also removed, accounting for $0.61\%$ of the data; this was achieved by training a new tokenizer with a vocabulary size of 52,000, derived from the BETO (Spanish BERT) pre-trained tokenizer \cite{CaneteCFP2020}.

\subsection{Post-OCR LLM Correction}

As previously discussed, LLMs have established a baseline for correcting OCR errors in historical texts \cite{thomas-etal-2024-leveraging, Post-OCR-Correction}. Detecting and fixing OCR errors from newspapers is challenging because these errors are often subtle and numerous. This problem is especially pronounced with 19th-century newspapers, where the quality of the paper and the outdated printing methods contribute to a high frequency of errors. These errors create significant noise and complicate the text correction process \cite{10.1145/1390749.1390753}.

In this paper, we use a technique for detecting OCR errors and correcting them using GPT-4o-mini and taking advantage of the fact that LLMs were trained mostly in modern language. This way, manually checked rules can classify corrections between errors, word surface forms, or none of both (hallucinations). These rules, explained in the following section, were revised and selected by a field expert who served as well as an evaluator for these corrections testing their precision for this case.

We employed a \textit{diff} algorithm to detect the differences between the original and corrected texts. This approach allowed us to fully leverage the LLM's ability to correct the text while ensuring a reliable and structured output. The \textit{diff} algorithm identifies added, removed, and changed parts between the two texts, similar to the functionality seen in GitHub's blame feature. By doing so, we can specify the exact changes made during the correction process, enabling us to classify these alterations effectively.

This method proved more effective than instructing the LLM to return corrections in a specific format, such as JSON, as the \textit{diff} algorithm produced shorter, more consistent, and less variable outputs. Additionally, this differentiation allows us to ignore any additions or deletions that result from LLM hallucinations, focusing instead on meaningful changes. An example of the original text, the corrected version, and the detected differences can be found in Appendix \ref{sec:appendix1}, as well as the parameters chosen for this step.

\subsection{Corrections Classification}

Once the corrections are detected and isolated through the \textit{diff} algorithm, the last step is to classify them. Still, first, it is important to state the main differences between the possible labels for each correction:
\begin{itemize}
    \item \textbf{Surface form:} In linguistics, the term \textit{surface form} (or word form) denotes the specific appearance of a word in a given context, contrasting with its lexical form, which pertains to its meaning \cite{sarveswaran-etal-2019-using}. During the 19th century in Latin America, certain words were documented with variant spellings reflecting language shifts over time. It's important to note that changes in surface forms do not necessarily alter the semantic content of the word, but rather represent orthographic modifications.
    \item \textbf{OCR error:} An OCR error, on the other hand, refers to every possible misread text from the real newspaper text. The OCR errors must be corrected but must be carefully separated newspaper linguistic "errors" that contribute to the linguistics of the time.
    \item \textbf{Hallucinations:} If none of the above is the case, the correction is an LLM hallucination or a translation to modern Spanish, which would be wrong, so these corrections must be omitted.
\end{itemize}

To enhance classification rule analysis, corrections were noted along with their frequency across the dataset to assess relevance. All corrections were converted to lowercase for effective grouping. Many corrections were reviewed and consolidated into a set of linguistic rules for categorization. This framework can be used to identify and analyze similar changes and classification rules in other languages and contexts. This paper presents a validated set of standardized rules and exceptions for classifying corrections in the LatamXIX dataset.

\subsubsection{Accent changes} Corrections involving only accent changes (addition or removal) between the original and corrected texts refer mostly to \textbf{surface forms}, given the differences between 19th-century Spanish accent rules and modern ones \cite{54f6385b-261f-3696-acfc-99604ae83a87}.  This includes varied accent expressions for the same word, such as "antes" sometimes written as "ántes". Surface forms pose problems for NLP tasks because, in Spanish, words without accents can have different meanings, such as "acepto" (present) and "aceptó" (past). Thus, for some NLP tasks, focusing on surface forms without accent changes may be preferable, which is another outcome presented in this paper.

\subsubsection{Specific changes} A set of letter-to-letter changes was extracted to represent key \textbf{surface words} and common \textbf{OCR errors}. For surface words, common changes include "y" for "i" or "g" for "j", e.g., "mui" for "muy" and "jeneral" for "general"; in fact, the connector "y" used to be written as "i" in most of the early 19th-century texts \cite{6490eab7-04fe-3604-aebf-873a615ed49e}. Common OCR errors include accent misreading or number confusion, such as "ó" read as "6" or "i" as "1". Appendix \ref{sec:appendix2} shows a list of surface form changes.

\subsubsection{Other letter-to-letter changes} When the number of letters in the original and corrected texts matches, changes generally refer to \textbf{OCR errors}, e.g., "la" misread as "In" or "señor" as "sefor".

\subsubsection{Remaining changes} Corrections not fitting the preceding categories are challenging to classify as OCR errors or hallucinations, particularly with multiword corrections. A text similarity ratio was computed based on positional character matches between the original and corrected texts. This ratio, combined with the number of words in the corrected text and correction frequency, helped categorize corrections. For instance, "ascripeión" to "suscripción" had a ratio of 0.76, while "que" to "como" had a ratio of 0.0, effectively distinguishing most cases.


\section{Results}

\begin{table}[t]
    \centering
    \begin{tabular}{|c|c|}
    \hline
    \textbf{Feature} & \textbf{Value}  \\ \hline
    Size             & $\sim 128 MB$ \\
    Rows             & $64,077$         \\
    Words           & $\sim 22 M$ \\
    Tokens           & $\sim 28.7 M$ \\
    Newspapers       & 197              \\
    Years Range       & 1806 - 1899              \\
    Total Corrections & 830,951 \\
    Surface Forms       & 37,492              \\
    Non-Accent Surface Forms       & 7,466              \\ 
    \% of OCR Error Corrections & 12.33\% \\ 
    \% of Hallucinations Detected & 77.96\% \\ \hline
    \end{tabular}
    \caption{\label{tab:overview} Final Historical Ink: LatamXIX LLM Post-OCR corrected dataset}
\end{table}

Following the outlined steps, we produced the LatamXIX dataset, as shown in Table \ref{tab:overview} and detailed in Appendix \ref{sec:appendix3}, alongside a flexible LLM OCR correction framework. This framework allows for easy interchange between datasets or LLMs, facilitating further research. We also compiled a list of 19th-century Latin American Spanish surface forms from newspapers and developed a general framework for detecting these forms in diverse contexts.

Old Spanish surface forms are particularly useful for semantic change detection, capturing meaning variations of specific words and aiding comparisons of their historical evolution across different periods and Spanish-speaking regions.

In terms of LLM post-OCR corrections, the system generated 830,951 corrections. However, a notable 78\% of these were classified as hallucinations, indicating the model's tendency to generate incorrect or fabricated content when uncertain. Only 12\% addressed actual OCR errors, reflecting the core objective of the framework. This gap highlights a key limitation of current LLM models in historical OCR correction, where distinguishing between genuine errors and hallucinations remains a challenge, especially in specialized datasets.

Moreover, due to Azure OpenAI's API content policy for the chosen LLM (GPT-4o-mini), 2,899 rows (4.52\%) were excluded from processing because they contained content flagged as harmful, violent, or sexual. This limitation underscores the challenges content moderation policies pose when applying LLMs to historical texts. The percentage of flagged content provides insight into the prevalence of such material in 19th-century Spanish, offering valuable perspectives for comparative analysis with modern Spanish \footnote{The dataset, surface forms, and processing steps are available in \url{https://github.com/historicalink/LatamXIX}}.

\section{Future Work}

While the OCR correction using LLMs has progressed towards a more automated pipeline, a substantial portion of rule definition within the presented framework still requires manual professional input. To advance this process, future work should aim to enhance the automation of the rule-defining procedures. By reducing the reliance on human expertise, we can improve both the efficiency and accuracy of the OCR correction framework.

\section{Limitations}

A significant limitation of this work is the reliance on manual evaluations for assessing OCR accuracy, as most evaluations and rule definitions were performed by experts. This manual process introduces subjectivity and limits scalability. The absence of a comprehensive automated evaluation method prevents more consistent accuracy assessments and restricts the ability to refine the framework based on objective metrics like Character Error Rate (CER).

\section{Acknowledgements}

We would like to thank the two anonymous reviewers from the EMNLP NLP4DH conference for their helpful feedback and suggestions.

\bibliography{acl_latex}

\appendix

\renewcommand\thefigure{\thesection\arabic{figure}}
\renewcommand\thetable{\thesection\arabic{table}}
\setcounter{figure}{0}    
\setcounter{table}{0}    

\section{LLM Correction}
\label{sec:appendix1}

\subsection{Prompt}

Below is the prompt used to request the LLM to correct the historical text extracted by the OCR model. This prompt remained unchanged for the correction of the entire dataset and was generated through manual trial and error, ensuring it was concise enough to accommodate the potential length of the text.

\begin{quote} 
\texttt{Dado el texto del siglo XIX entre \`{}\`{}\`{}, retorna únicamente el texto corrigiendo los errores ortográficos sin cambiar la gramática. No corrijas la ortografía de nombres:}

\texttt{\`{}\`{}\`{}}

\texttt{\{text\}}

\texttt{\`{}\`{}\`{}}

\end{quote} 
\leavevmode \newline
Equivalent to the following prompt in English:

\begin{quote} 
\texttt{Given the 19th-century text between \`{}\`{}\`{}, return only the text with spelling errors corrected without changing the grammar. Do not correct the spelling of names:}

\texttt{\`{}\`{}\`{}}

\texttt{\{text\}}

\texttt{\`{}\`{}\`{}}

\end{quote} 

\subsection{Example}

The LLM response was successful for most of the texts except for some cases where Azure's Content Policy was triggered due to text content, and for very long texts where the model started to hallucinate the whole text. An example of an original text, its retrieved LLM correction, and all the changes detected by the diff algorithm is the following (\textcolor{blue}{surface forms} and \textcolor{red}{OCR errors}) is:

\begin{itemize}
    \item \textbf{Original:} La publicacion del Oso se harà dos veces cada se mana, y constará de un pliego en cuarto ; ofreciendo à mas sus redactores, dar los gravados oportunos, siempre que loexija el asuntode que trate. Redactado por un Num. 8. TEMA del Periodico. POLITICA MILITAR. OCTAVA SESION. Abierta la sesion á las dore y un minuto de la noche , 25 de Febrero de 1845 , con asistencia de todos los Señores Representantes, se leyó y aprobó la acta de la Asamblea anterior , ménos en lo tocante à la torre del Convento de Santo Domingo, punto que quedó para ventilarse en mejor ocasion. En seguida se dió cuenta de una nota del Ejecutivo , referente à que urjía la necesidad de organizar un Ejército ; pues decia el Excmo. Decano: - "Un poder sin bayonetas vale tanto como un cero puesto á la izquierda." 
    \item \textbf{Corrected:} La \textcolor{blue}{publicación} del Oso se \textcolor{blue}{hará} dos veces cada \textcolor{red}{semana}, y constará de un pliego en cuarto; ofreciendo \textcolor{red}{además} sus redactores, dar los \textcolor{blue}{grabados} oportunos, siempre que \textcolor{red}{lo exija} el \textcolor{red}{asunto de} que trate. Redactado por un Num. 8. TEMA del \textcolor{blue}{Periódico}. \textcolor{blue}{POLÍTICA} MILITAR. OCTAVA \textcolor{blue}{SESIÓN}. Abierta la \textcolor{blue}{sesión a} las \textcolor{red}{dos} y un minuto de la noche, 25 de Febrero de 1845, con asistencia de todos los Señores Representantes, se leyó y aprobó la acta de la Asamblea anterior, \textcolor{blue}{menos} en lo tocante \textcolor{blue}{a} la torre del Convento de Santo Domingo, punto que quedó para ventilarse en mejor \textcolor{blue}{ocasión}. \textcolor{red}{Enseguida} se \textcolor{blue}{dio} cuenta de una nota del Ejecutivo, referente \textcolor{blue}{a} que \textcolor{blue}{urgía} la necesidad de organizar un Ejército; pues \textcolor{blue}{decía} el Excmo. Decano: - "Un poder sin bayonetas vale tanto como un cero puesto \textcolor{blue}{a} la izquierda."
\end{itemize}

\section{Specific Surface Form Changes}
\label{sec:appendix2}
For the surface form extraction from the texts and its differentiation from OCR errors and LLM hallucinations, a set of surface form changes was constructed for 19th-century Latin American Spanish. The complete set of known changes with an example for each case is presented in Table \ref{tab:sf-changes}.

\begin{table}[h!]
  \centering
\begin{tabular}{|c|c|}
    \hline
    \textbf{Change} & \textbf{Example} \\
    \hline
    á $\leftrightarrow$ a & {hara $\rightarrow$ hará} \\
    é $\leftrightarrow$ e & {fué $\rightarrow$ fue} \\
    í $\leftrightarrow$ i & {decia $\rightarrow$ decía} \\
    ó $\leftrightarrow$ o & {ocasion $\rightarrow$ ocasión} \\
    ú $\leftrightarrow$ u & {ningun $\rightarrow$ ningún} \\
    i $\leftrightarrow$ y & {mui $\rightarrow$ muy} \\
    j $\leftrightarrow$ g & {jente $\rightarrow$ gente} \\
    v $\leftrightarrow$ b & {gravado $\rightarrow$ grabado} \\
    s $\leftrightarrow$ x & {espiró $\rightarrow$ expiró} \\
    j $\leftrightarrow$ x & {méjico $\rightarrow$ méxico} \\
    c $\leftrightarrow$ s & {faces $\rightarrow$ fases} \\
    s $\leftrightarrow$ z & {dies $\rightarrow$ diez} \\
    z $\rightarrow$ c & {doze $\rightarrow$ doce} \\
    q $\rightarrow$ c & {quatro $\rightarrow$ cuatro} \\
    n $\rightarrow$ ñ & {senor $\rightarrow$ señor} \\
    ni $\rightarrow$ ñ & {senior $\rightarrow$ señor} \\
    k $\rightarrow$ qu & {nikel $\rightarrow$ níquel} \\
    k $\rightarrow$ c & {kiosko $\rightarrow$ quiosco} \\
    ou $\rightarrow$ u & {boulevar $\rightarrow$ bulevar} \\ 
    s $\rightarrow$ bs & {suscriciones $\rightarrow$ subscripciones} \\
    c $\rightarrow$ pc & {suscriciones $\rightarrow$ subscripciones} \\
    s $\rightarrow$ ns & {trasportar $\rightarrow$ transportar} \\
    t $\rightarrow$ pt & {setiembre $\rightarrow$ septiembre} \\
    rt $\rightarrow$ r & {libertar $\rightarrow$ liberar} \\
    r $\leftrightarrow$ rr & {vireinato $\rightarrow$ virreinato} \\
    ...lo $\rightarrow$ lo ... & {cambiólo $\rightarrow$ lo cambió} \\
    ...se $\rightarrow$ se ... & {acercóse $\rightarrow$ se acercó} \\
    \hline
\end{tabular}
  \caption{Set of Surface Form change rules to extract them from the LatamXIX dataset}
  \label{tab:sf-changes}
\end{table}

\ 
\newpage

\section{Dataset Overview}
\label{sec:appendix3}
\setcounter{figure}{0}    
\setcounter{table}{0}    

\begin{figure}[h!]
  \centering
  \includegraphics[width=\linewidth]{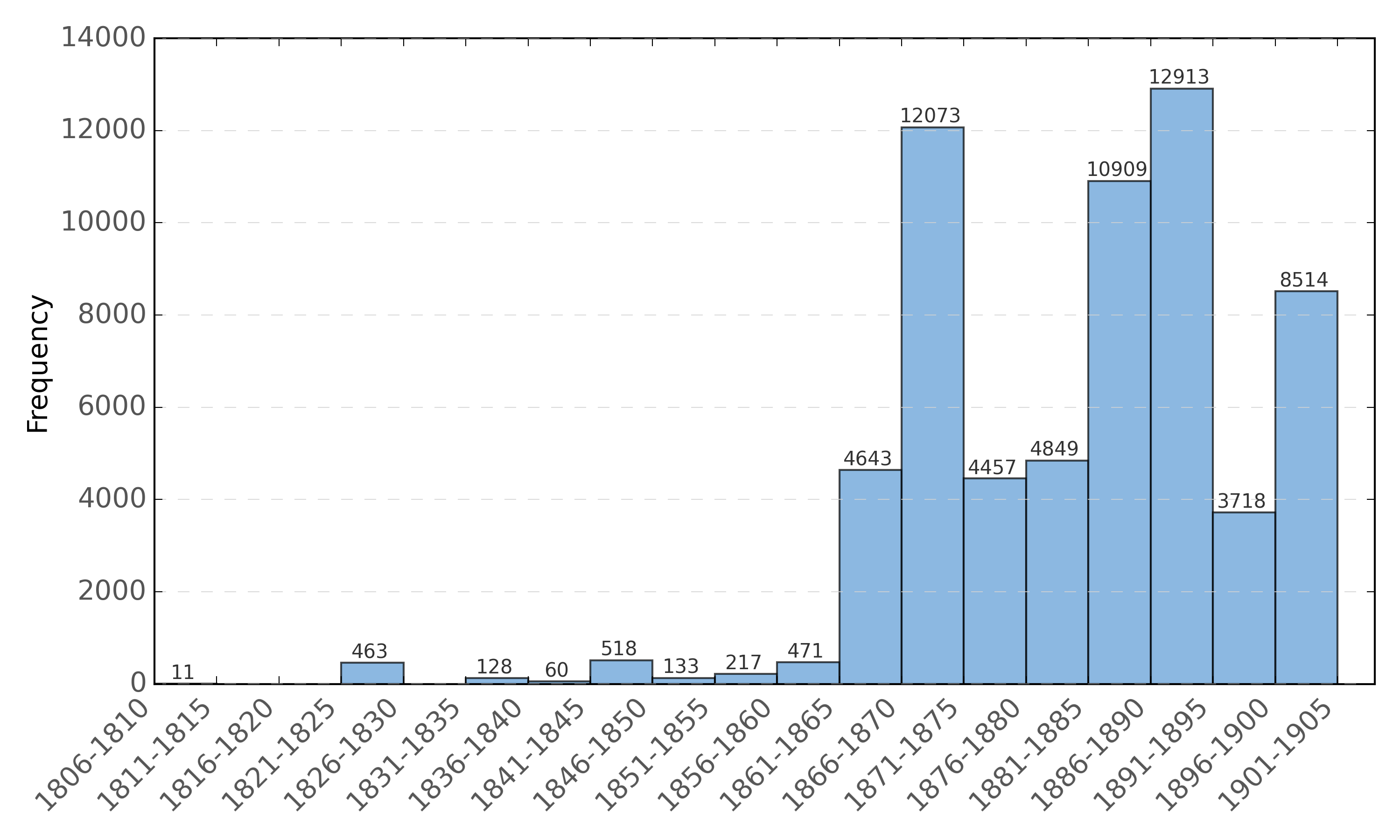}
  \caption{\label{fig:overview} LatamXIX dataset decade distribution}
\end{figure}

A more specific overview of the dataset is described in Figure \ref{fig:overview} and Table \ref{tab:overview-country}.

\begin{table}[h!]
    \centering
    \begin{tabular}{|c|c|}
    \hline
    \textbf{Country} & \textbf{Presence (\%)}  \\ \hline
    Mexico & 49.59\% \\ 
    Argentina & 21.23\% \\ 
    Colombia & 12.53\% \\ 
    Peru & 8.43\% \\ 
    Chile & 6.39\% \\ 
    Panama & 0.83\% \\ 
    Venezuela & 0.52\% \\ 
    Uruguay & 0.17\% \\ 
    France & 0.16\% \\ 
    Ecuador & 0.09\% \\ 
    Spain & 0.06\% \\ 
    \hline
    \end{tabular}
    \caption{\label{tab:overview-country} LatamXIX dataset presence distribution grouped by country}
\end{table}

\end{document}